\documentclass[journal,twoside,web]{ieeecolor}

\usepackage{etoolbox}
\makeatletter
\@ifundefined{color@begingroup}%
{\let\color@begingroup\relax
\let\color@endgroup\relax}{}%
\def\fix@ieeecolor@hbox#1{%
\hbox{\color@begingroup#1\color@endgroup}}
\patchcmd\@makecaption{\hbox}{\fix@ieeecolor@hbox}{}{\FAILED}
\patchcmd\@makecaption{\hbox}{\fix@ieeecolor@hbox}{}{\FAILED}

\usepackage{tmi}
\usepackage{cite}
\usepackage{amsmath,amssymb,amsfonts}
\usepackage{algorithmic}
\usepackage{graphicx}
\usepackage{textcomp}
\usepackage{array} 
\usepackage{multirow} 
\usepackage[linesnumbered,ruled,lined]{algorithm2e} 
\def\BibTeX{{\rm B\kern-.05em{\sc i\kern-.025em b}\kern-.08em
    T\kern-.1667em\lower.7ex\hbox{E}\kern-.125emX}}
\begin{document}
\title{DFGET: Displacement-Field Assisted Graph Energy Transmitter for Gland Instance Segmentation}
\author{Caiqing Jian, Yongbin Qin, and Lihui Wang
\thanks{This work was partially funded by the National Natural Science Foundation of China (Grant No. 62066008 and 62161004), 
Guizhou Provincial Science and Technology Projects (QianKeHe ZK [2021] Key 002, QianKeHe[2020]1Y255). 
(Corresponding author: Yongbin Qin). }
\thanks{Caiqing Jian, Yongbin Qin, and Lihui Wang are with the Engineering Research Center of 
Text Computing \& Cognitive Intelligence, Ministry of Education, Key Laboratory of Intelligent 
Medical Image Analysis and Precise Diagnosis of Guizhou Province, State Key Laboratory of Public 
Big Data, College of Computer Science and Technology, Guizhou University, Guiyang 550025, China 
(e-mail: gs.cqjian21@gzu.edu.cn; ybqin@gzu.edu.cn; wlh1984@gmail.com).}
\thanks{This work has been submitted to the IEEE for possible publication. Copyright may be transferred without notice, after which this version may no longer be accessible.}
}
\maketitle 

\begin{abstract}
    Gland instance segmentation is an essential but challenging task in the diagnosis and treatment of adenocarcinoma. The existing models usually achieve gland instance segmentation through multi-task learning and boundary loss constraint. However, how to deal with the problems of gland adhesion and inaccurate boundary in segmenting the complex samples remains a challenge. In this work, we propose a displacement-field assisted graph energy transmitter (DFGET) framework to solve these problems. Specifically, a novel message passing manner based on anisotropic diffusion is developed to update the node features, which can distinguish the isomorphic graphs and improve the expressivity of graph nodes for complex samples. Using such graph framework, the gland semantic segmentation map and the displacement field (DF) of the graph nodes are estimated with two graph network branches. With the constraint of DF, a graph cluster module based on diffusion theory is presented to improve the intra-class feature consistency and inter-class feature discrepancy, as well as to separate the adherent glands from the semantic segmentation maps. Extensive comparison and ablation experiments on the GlaS dataset demonstrate the superiority of DFGET and effectiveness of the proposed anisotropic message passing manner and clustering method. Compared to the best comparative model, DFGET increases the object-Dice and object-F1 score by 2.5\% and 3.4\% respectively, while decreases the object-HD by 32.4\%, achieving state-of-the-art performance.
\end{abstract}


\section{Introduction}
\label{sec:introduction}
\IEEEPARstart{A}{ccurate} and consistent gland instance segmentation plays an important role in the grading of cancers originating from glandular tissues. However, manual segmentation is not only time-consuming and labor-intensive, but also suffers from inconsistency in segmentation criteria across physicians, leading to inconsistency in cancer ratings\cite{b1}-\cite{b3}. In the early stages of the development of gland segmentation algorithms, researchers used image processing algorithms such as region growing \cite{b4},\cite{b5}, and watershed algorithm \cite{b6}, \cite{b7} to segment gland instances. However, such traditional methods can only handle glands with regular morphology and clear boundaries. With the successful application of deep learning in the field of natural image and medical image segmentation \cite{b8}-\cite{b14}, more and more works have started to use deep models for gland instance segmentation.

Since the 2015 MICCAI Gland Segmentation Challenge (GlaS), numerous approaches based on convolutional neural networks have emerged. For instance, Chen \emph{et al.} \cite{b15} proposed a multi-task model named DCAN, which integrates gland semantic segmentation and contour segmentation tasks to predict the instance maps. Inspired by this work, many methods based on multi-task learning framework have been proposed to address the issue of gland adhesion \cite{b16}-\cite{b18}. In addition to multi-task networks, some meaningful attempts in learning strategies and loss functions have also been presented. For example, Yan \emph{et al.} \cite{b19}, \cite{b20} transformed the segmentation task into curve fitting of segments to solve the problem of boundary uncertainty caused by inconsistent manual labeling, but this segment-by-segment fitting strategy is very time-consuming. Graham \emph{et al.} \cite{b21} introduced multiple image transformations and averaged their predictions as a refined segmentation result to exclude uncertain regions. To further refine the segmentation, Gunesli \emph{et al.} \cite{b22} proposed a staged refinement strategy where the loss weight of a pixel is adjusted according to its accuracy in the previous predictions. Xie \emph{et al.} \cite{b23}, \cite{b24} proposed an intra-pair and inter-pair consistency (I$^2$CS) module to enhance the semantic feature representation ability. They also designed an object-level loss to address adherent glands. The above methods have alleviated the problems of adhesion and inaccurate segmentation to some extent, but their performance in dealing with the complex samples is still not satisfactory. 
In recent years, graph neural network (GNN), represented by message-passing mechanisms, has made significant theoretical and methodological developments \cite{b25}-\cite{b29}. Numerous informative works have emerged on medical image segmentation using GNNs to learn shape prior and topological associations. For instance, Meng \emph{et al.}\cite{b30} used a signed distance function to divide the pixels into region and boundary zones, and then developed graph reasoning modules to aggregate and link the region and boundary node features respectively. Through the interactions of different graph reasoning modules, the segmentation performance can be promoted. Yao \emph{et al.} \cite{b31} used a graph network to learn the shape prior of organs, in which the grid coordinates and the corresponding features are updated by the graph to generate the shape-aware constraints. Zhao \emph{et al.} \cite{b32} pointed out that concatenating simply the coordinates and features may confuse the model. Accordingly, they proposed a two-stream graph convolutional network to process visual features and coordinates separately, and the informations in two streams are exchanged by attention modules. Liu \emph{et al.} \cite{b33} proposed an interactive graph network (IGNet) which transfers the messages about the domain-common prototypes with a graph to promote the semi-supervised segmentation performance. From the above literatures, we can see that, the message passing mechanism in the graph can enhance the interaction between different regions. Therefore, in certain cases, using graph models to segment may achieve better performance.

Currently, the graph attention network (GAT)\cite{b26} and its variants are commonly regarded as the benchmark for graph models. However, their static attention severely restricts the expressiveness. Despite attempts to enhance them with dynamic attention \cite{b29} or devising novel GNN models based on physical principles \cite{b34}-\cite{b39}, they are still upper-bounded by Weisfeiler-Lehman test\cite{b27}. Such limitation makes them unable to distinguish isomorphic graphs \cite{b27}, which is not conducive to segmenting complex adenoma instances. To address this issue, inspired by the anisotropic diffusion of energy, we propose a displacement-field assisted graph energy transmitter (DFGET) framework, which incorporates a novel message passing manner to enable the model to distinguish the isomorphic graphs, and a diffusion displacement field aided clustering method to separate the adherent glands. With the help of these two core components, DFGET can effectively improve the segmentation performance, especially for the complex samples with adherent instances and irregular boundaries. 

\begin{figure*}[t] %
    \centering
    \includegraphics[width=0.8\textwidth]{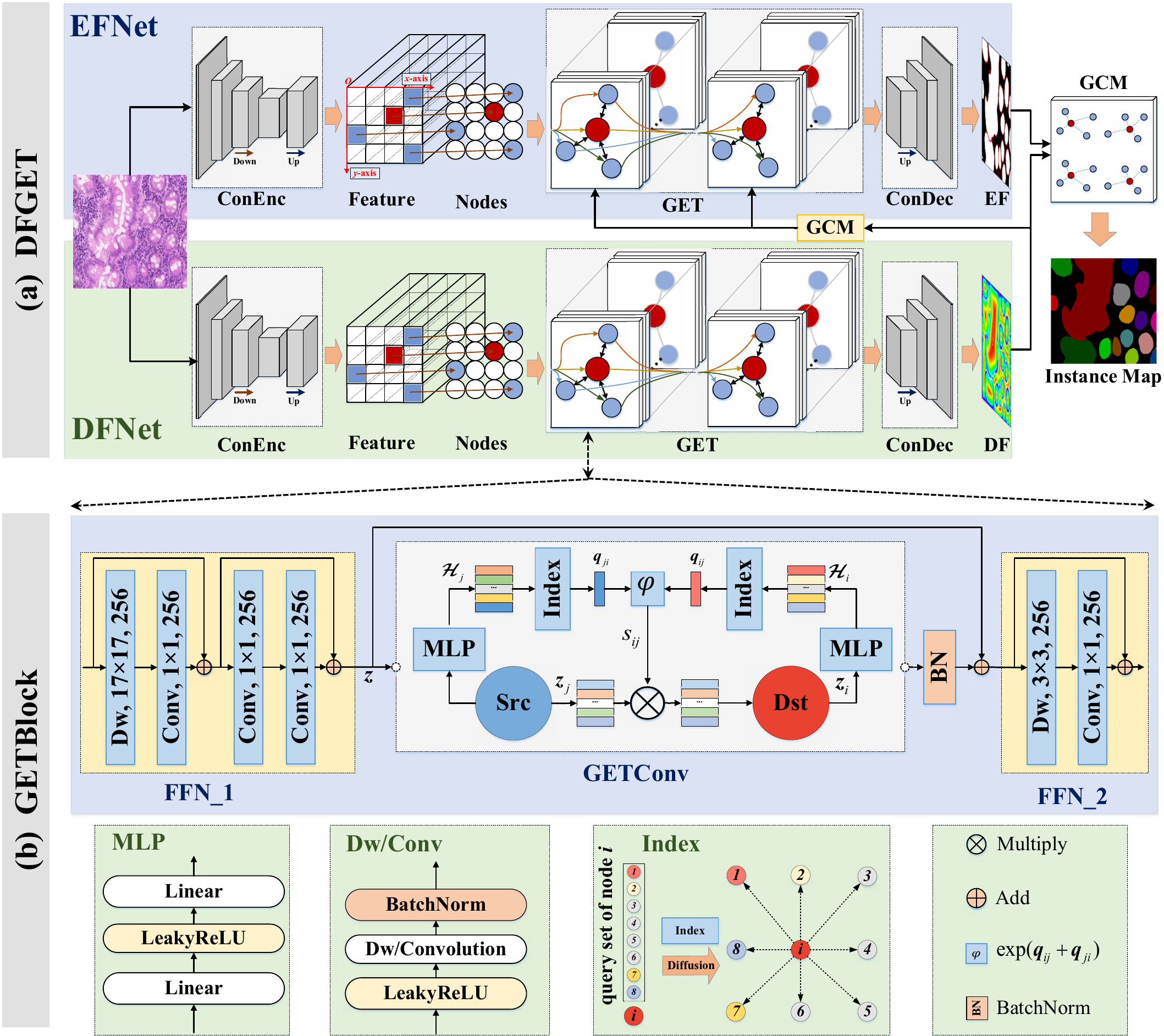}
    \caption{
        \textbf{(a) Flowchart of DFGET}. Input image ${\boldsymbol{I}} \in {\mathbb{R}^{256 \times 384 \times 3}}$ 
        is firstly fed to DFNet for displacement field (DF, $\hat{\textbf{\textit{F}}}$ for simplicity), 
        and then to EFNet for the energy field (EF, $\hat{\textbf{\textit{E}}}$ for simplicity). The output ${\textbf{\textit{Feature}}} \in {\mathbb{R}^{64 \times 96 \times 256}}$
        of ConEnc is assigned to 6144 nodes of the graph (only 16 nodes are shown for simplicity), and each node is characterized by 256 
        dimensions. GET performs anisotropic message passing, with the output having the same dimensionality as the input. There are 
        two GCMs, one of which accepts the displacement field of DFNet and obtains the node clustering to constrain the message passing 
        of GETBlock in EFNet. The other GCM is responsible for integrating the outputs of DFNet and EFNet to obtain the instance map. \ \ \ 
        \textbf{(b) Architecture of GETBlock.} 17$\times$17 or 3$\times$3 and number 256 correspond to kernel size and output 
        channel in depthwise convolution; [Conv,1$\times$1,256] means pointwise convolution with output 
        channel 256; Src and Dst represent the source node and target node, respectively. For 
        simplicity, only one target node and one source node are shown here. In the actual 
        message graph (MG), each node can be a target node, and each target node has multiple 
        source nodes; The Index module generates query messages from $\mathcal{H}$ by indexing 
        specific channels of $\mathcal{H}$, and the indices are defined by the spatial locations of the 
        neighboring nodes. All network layers (including convolution, BN, and MLP) are 
        shared among nodes.
        }
    \label{fig1}
    \vspace{-10pt} 
\end{figure*}

\section{Methods}
The detailed architecture of DFGET is shown in Fig. \ref{fig1}, which includes three components: the displacement field network (DFNet), the energy field network (EFNet), and the Graph Cluster Module (GCM). The displacement field output by DFNet has two roles: one is to constrain the message passing of GET in EFNet, and the other is to cluster nodes in GCM. EFNet is responsible for predicting the energy field (semantic segmentation map). GCM converts the displacement field and energy field into an instance map. The subsequent sub-sections will detail the principles and technical details of these core modules of DFGET.

\subsection{EFNet for Semantic Segmentation Based on Anisotropic Diffusion}

As shown in Fig. \ref{fig1} (a), in EFNet, the input image passes through firstly a convolutional encoder (ConEnc) to extract the feature maps; Then, each pixel in the feature map is taken as a graph node, and a graph model named GET is used to further refine the feature map; Finally, a convolutional decoder (ConDec) is used to upsample the refined feature map and output the energy map $\hat{\textbf{\textit{E}}}$, namely the semantic segmentation result. Note that, the graph model used in EFNet is totally different from the existing ones. To deal with the limits of the existing GNNs in distinguishing isomorphic graphs, and to further refine the semantic features of nodes, we proposed a graph energy transmitter (GET) model based on diffusion theory. 

It is well known that the diffusion equation can be expressed as:
\begin{equation} \label{eq1}
\begin{aligned} 
\frac{\partial \boldsymbol{Z}^{(t)}}{\partial t} &=\nabla^*\left(\boldsymbol{S}\left(\boldsymbol{Z}^{(t)}, t\right) \odot \nabla \boldsymbol{Z}^{(t)}\right),
\end{aligned}
\end{equation}
where ${{\boldsymbol{Z}}^{(t)}}$ is the energy at time \emph{t};
${\nabla }^{*}$, $\nabla$, and ${\boldsymbol{S}}({{\boldsymbol{Z}}^{(t)}},t)$ represent divergence operator, gradient operator, and diffusivity, respectively; $\odot$ is Hadamard product. On a discrete particle system, we introduce time step $\tau$ and use numerical finite difference to expand the diffusion process in (\ref{eq1}) into an iterative form. We then shift the terms to obtain a generalized energy transmitting equation:
\begin{equation} \label{eq2}
\boldsymbol{z}_i^{(t+1)}=\left(1-\tau \sum_{j=1}^N s_{i j}^{(t)}\right)
\boldsymbol{z}_i^{(t)}+\tau \sum_{j=1}^N s_{i j}^{(t)} \boldsymbol{z}_j^{(t)}, 
\end{equation}
where $\boldsymbol{z}_i^{(t)}$ is the energy of particle $i$ after the $t$-th iteration, and $s_{i j}^{(t)}$ represents the diffusivity between particle $i$ and its neighbor $j$. We normalize the sum of neighborhood diffusivity to 1 in (\ref{eq2}) and restrict the neighborhood to $\mathcal{N}_i$, resulting in:
\begin{equation} \label{eq3}
\boldsymbol{z}_i^{(t+1)}=\left(1-\tau \right)
\boldsymbol{z}_i^{(t)}+\tau \sum_{j \in \mathcal{N}_i} s_{i j}^{(t)} \boldsymbol{z}_j^{(t)}.
\end{equation}

To omit the hyperparameter $\tau$, (\ref{eq3}) can be reformulated  as (\ref{eq4}), which is similar to the message passing in the graph. 
\begin{equation} \label{eq4}
\boldsymbol{z}_i^{(t+1)} = \boldsymbol{z}_i^{(t)} + 
\operatorname{BN} \left(\sum_{j \in \mathcal{N}_i} s_{i j}^{(t)} \boldsymbol{z}_j^{(t)}\right)
\end{equation}
That means, the energy $\boldsymbol{z}_i$ of the $i$-th particle can be analogized to the semantic features of node $i$, and the diffusivity $s_{ij}$ can be taken as the weight for information exchange between nodes $i$ and $j$ in a graph. In the existing GNNs, the query message of node $i$ and the key message of node $j$ are isotropically passed to all their neighbor nodes, no matter where the nodes locate. We call this message passing manner as isotropic diffusion, which cannot distinguish isomorphic graphs. To deal with this issue, we introduced a novel message passing method based on anisotropic diffusion, in which the diffusivity $s_{ij}$ is calculated with GETConv in Fig. \ref{fig1} (b). Specifically, the features of nodes $\boldsymbol{Z}^{(t)}=\left[\boldsymbol{z}_i^{(t)}\right]_{i=1}^N$ pass through a MLP to obtain a query message set $\boldsymbol{\mathcal{H}}^{(t)} \in{R^{N\times n}}$, $n$ is the maximum number of neighbors for any node. 
\begin{equation} \label{eq5}
    \boldsymbol{\mathcal{H}}^{(t)}=\operatorname{MLP^{(t)}}\left(\boldsymbol{Z}^{(t)}\right).
\end{equation}

Using index operation through both channel and node dimensions, the queries for nodes $i$ and $j$ can be obtained,
 \begin{equation} \label{eq6}
 \begin {aligned}
    \boldsymbol{q}_{ij}^{(t)} &= \boldsymbol{\mathcal{H}}^{(t)} \left\{i, c_j\right\} 
    = \boldsymbol{\mathcal{H}}_{ic_j}, \\
    \boldsymbol{q}_{ji}^{(t)} &= \boldsymbol{\mathcal{H}}^{(t)} \left\{j, c_i\right\} 
    = \boldsymbol{\mathcal{H}}_{jc_i},
 \end {aligned}
\end{equation}
where $\mathcal{H}_{ic_j}$ denotes the query message passed from node $i$ to its $c_j$-th neighbor, while $\mathcal{H}_{jc_i}$ indicates the query message passed from node $j$ to its $c_i$-th neighbor, with $c_j, c_i \in {[0, n-1]}$ being the neighbor indices of nodes $i$ and $j$.

Accordingly, the diffusivity between nodes $i$ and $j$ is defined as:
\begin{equation} \label{eq7}
    s_{ij}^{(t)}=\exp \left(\boldsymbol{q}_{ij}^{(t)}+\boldsymbol{q}_{ji}^{(t)}\right),
\end{equation}
From (\ref{eq6}) and (\ref{eq7}), we notice that if the indices of the neighbors are different, the messages passed or received by the same node are anisotropic, which enables the model to possess the ability to distinguish isomorphic graphs. By utilizing the diffusivity in (\ref{eq7}) to update the graph features, we can enhance the node's characteristics. Consequently, this advancement contributes to improving the accuracy of segmentation.

\subsection{Instance Segmentation Based on DFNet and GCM}
To address the issue of gland adhesion, this work introduces a DFNet to infer the displacement field ($\hat{\textbf{\textit{F}}}$) of graph nodes, which estimates the motion direction of each node relative to its corresponding instance center. With the constraints of displacement field and the post processing of GCM, the instance segmentation results without adhesion can be finally derived. 

As shown in Fig. \ref{fig1} (a), the network structure of DFNet is similar to that of EFNet, but their objective is totally different and they do not share parameters. The EFNet is used for semantic segmentation, its learning target is easy to derive from the instance label, whereas the expected output of DFNet ($\hat{\textbf{\textit{F}}}$) is not intuitive to calculate. For a given instance label map, it can be expressed as a graph, with each pixel being a node. The cartesian coordinates of the pixels along $x$-axis and $y$-axis act as the initial node features, noted as $\textbf{\textit{D}} ^{(0)} \in {\mathbb{R}^{hw \times 2}}$, where $h$ and $w$ denote the height and width of the image. According to the graph diffusion theory in (\ref{eq2}) and the instance label of each node, we update the features (coordinates) of each node with (\ref{eq8}), where $L_{ins}^i$ and $L_{ins}^j$ denote the instance labels of nodes $i$ and $j$:

\begin{equation} \label{eq8}
\begin{gathered}
    \textbf{\textit{D}}_i^{(t)}=\frac{1}{\sum_{j \in \mathcal{N}_i} s_{i j}^{(t-1)}} \sum_{j \in \mathcal{N}_i} s_{i j}^{(t-1)} \textbf{\textit{D}}_j^{(t-1)} \\
    \text { s.t. } \tau =1 / \sum_{j \in \mathcal{N}_i} s_{i j}^{(t-1)}, \ \ s_{i j}^{(t-1)}=\left\{\begin{array}{ll}
    1, & \text{if } L_{ins}^i = L_{ins}^j\\
    0, & \text{else} 
\end{array} .\right.
\end{gathered}
\end{equation}

Iterating $t$ times according to (\ref{eq8}), the coordinates of the nodes in the same gland instance will go towards the center of this instance. Accordingly, the adherent nodes at the boundaries will be separated. The corresponding ground truth displacement field $\textbf{\textit{F}}$ is expressed as: 
\begin{equation} \label{eq9}
    \textbf{\textit{F}} = \textbf{\textit{D}} ^{(t)} - \textbf{\textit{D}} ^{(0)}.
\end{equation}
By minimizing the difference between $\textbf{\textit{F}}$ and $\hat{\textbf{\textit{F}}}$, the parameters of DFNet can be optimized. 

As we mentioned above, the estimated displacement field $\hat{\textbf{\textit{F}}}$ can make the nodes move towards its instance center, therefore it has potential to improve the feature consistency of nodes in the same cluster and to separate the adherent glands. To this end, we propose GCM. As illustrated in Fig. \ref{fig1} (a), GCM is firstly embedded in the EFNet to promote the intra-class feature consistency and inter-class discrepancy, and then used to separate the gland instance based on the semantic segmentation map output by EFNet and the displacement map $\hat{\textbf{\textit{F}}}$ output by DFNet. In the following paragraphs, the basic idea of GCM will be detailed.

According to the coordinates  $\textbf{\textit{D}}$ and the displacement field $\hat{\textbf{\textit{F}}}$ of all the pixels, a directional transmitting graph $tg$ is constructed, with the indices of pixels representing the graph nodes $\mathcal{V}$, the energy $\hat{\textbf{\textit{E}}}$ being the message of the nodes, and the displacement vector originating from any node $i$ as its edge $\mathcal{E}_i$, namely:

\begin{equation} \label{eq10}
\begin{aligned}
tg&=\mathcal{G}(\textbf{\textit{D}}, \hat{\textbf{\textit{F}}})= \left\{ (\mathcal{V}_i, \mathcal{E}_i), \ \ i = 1, 2,..., N  \right\} \\
\text { s.t. } \mathcal{V}_i&=\operatorname{nid}\left(\textbf{\textit{D}}_i\right),\\ 
               \mathcal{E}_i&=\operatorname{nid}\left(\textbf{\textit{D}}_i\right) \rightarrow 
\operatorname{nid}\left(\textbf{\textit{D}}_i+ \hat{\textbf{\textit{F}}}_i\right),\\
tg&.\textbf{\textit{mes}}_i = \hat{\textbf{\textit{E}}}_i, \\
\end{aligned}
\end{equation}
where $\rightarrow$ is a directed edge from $\operatorname{nid}\left(\textbf{\textit{D}}_i\right)$ to 
$\operatorname{nid}\left(\textbf{\textit{D}}_i+ \hat{\textbf{\textit{F}}}_i\right)$, and $\operatorname{nid}\left(\cdot\right) $ is used to calculate the id of the node based on the pixel coordinates, defined by:

\begin{equation} \label{eq11}
\operatorname{nid}\left(\textbf{X}\right) = \textbf{X}_x * w + \textbf{X}_y
\end{equation} 
where $\textbf{X}$ represents the coordinates of any node with $\textbf{X}_x$ and $\textbf{X}_y$ being the coordinates along $x$-axis and $y$-axis, respectively; $w$ is the image width.

\begin{figure}[t] 
    \vspace{-10pt}
    \centering
    \includegraphics[width=0.49\textwidth]{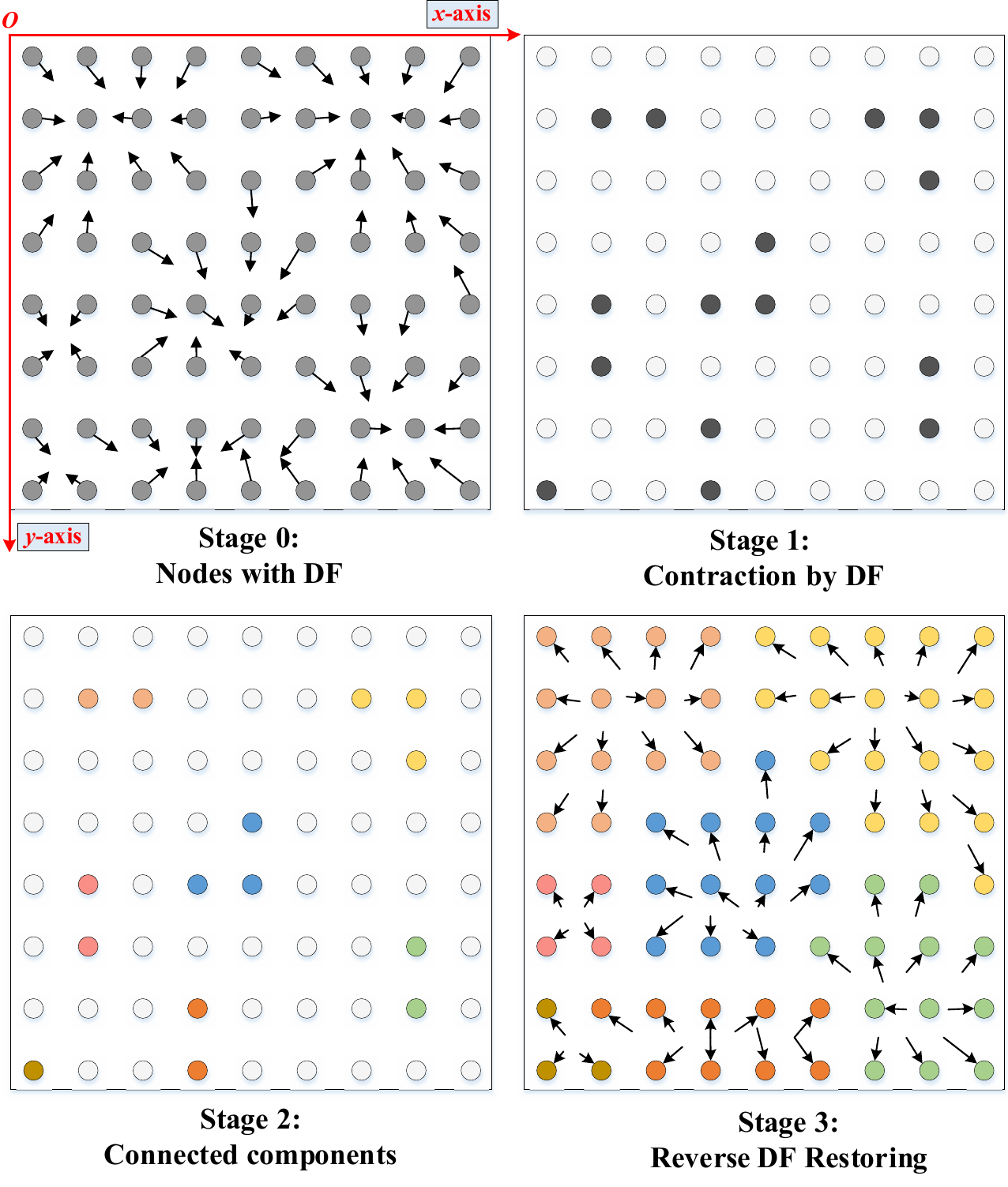}
    \caption{
    Schematic diagram of GCM. The grid coordinate system takes the upper left corner of the picture as the origin, 
    with the height direction as the $y$-axis and the width direction as the $x$-axis. The process of GCM consists of 
    three stages: 1) The contraction of nodes (pixels) driven by DF is achieved by (\ref{eq12}) on $tg$; 
    2) Connected component detection is performed on the clustered nodes, different instance IDs (different colors in "Stage 2)" are assigned 
    to different clusters; 3) Through message passing on $rg$, the instance IDs are reversely passed to the surrounding nodes, thus restoring the complete instances.
    }
    \label{fig2}
    \vspace{-10pt} 
\end{figure}

Along the directional edges, the message of each node in $tg$ is updated with:
\begin{equation} \label{eq12}
{tg.\textbf{\textit{mes}}}_i^{(t+1)}=\sum_{j \in \mathcal{N}_i} tg.\textbf{\textit{mes}}_j^{(t)}.
\end{equation}
where $\mathcal{N}_i$ includes all the nodes pointing to the node $i$. The displacements point to the centers of corresponding instances, which results in zero in-degree 
of boundary nodes on $tg$. Therefore, after iterations using (\ref{eq12}), the message of boundary 
nodes on $tg$ becomes zero while the message of internal nodes accumulates, which is equivalent to the fact that the nodes at boundaries of instances contract towards the interior of the instances as shown in Fig. \ref{fig2}. After that, the connected component detection is implemented on the contracted graph message map to generate the initial instance or cluster label $\textbf{\textit{Ins}}_i$ for each node $i$:
\begin{equation} \label{eq13}
\textbf{\textit{Ins}}_i=\left\{\begin{array}{cc}
0 & \text { if } tg.\textbf{\textit{mes}}_i=0 \\
int & \text { else }
\end{array}\right.
\end{equation}
where ${int}$ is an integer from 1 to the number of instances. To determine which cluster or instance that each boundary node belongs to, we construct a reverse graph $rg$:
\begin{equation} \label{eq14}
rg = \operatorname{reverse}\left( tg \right),
\end{equation}
$rg$ shares nodes with $tg$, but reversing the direction of edges. The message of each node in $rg$ is initialized by their instance label, i.e., $rg.\textbf{\textit{mes}}_i=\textbf{\textit{Ins}}_i$. By performing message passing on $rg$ similar to (\ref{eq12}), the instance labels for boundary nodes can be determined, i.e., the final instance map $\textbf{\textit{InsMap}}=rg.\textbf{\textit{mes}}$. The detailed process of GCM is given in \textbf{Algorithm 1}.

\begin{algorithm}[t]
    \begin{center}
    \begin{minipage}{0.49\textwidth} 
        \IncMargin{0.25em}
            \DontPrintSemicolon
            \SetKwInOut{Input}{Input}
            \SetKwInOut{Output}{Output}
                
            \caption{Graph Cluster Module}
            \label{GCM}
            \Input{\textbf{\textit{D}}, $\hat{\textbf{\textit{E}}}$, $\hat{\textbf{\textit{F}}}$}
            \Output{Instance Map \textbf{\textit{InMap}}}
            /* stage 1: Contraction of nodes driven by $\hat{\textbf{\textit{F}}}$ */ \;
            \ $tg = \mathcal{G}(\textbf{\textit{D}}, \hat{\textbf{\textit{F}}})$ \  \;
            \ $tg.\textbf{\textit{mes}} = \hat{\textbf{\textit{E}}}$ \ 
             \; 
            \For{$t \leftarrow 1$ \KwTo $T_0$}{   
                \ $tg.\textbf{\textit{mes}}_i$ $\leftarrow$ $\sum\limits_{j \in {{\cal N}_i}} 
                {tg.\textbf{\textit{mes}}_j}, \ \forall \ i \in tg$ \; 
                }
            /* stage 2: Connected component detection */ \;
            \ $\textbf{\textit{Ins}} = \operatorname{connectedComponents}
            (tg.\textbf{\textit{mes}})$ \;
            /* stage 3: Reverse graph to complete instance labels */ \;
            \ $rg = \operatorname{reverse}(tg)$ \;
            \ $rg.\textbf{\textit{mes}} = \textbf{\textit{Ins}}$ \;
            \For{$t \leftarrow 1$ \KwTo $T_1$}{   
                \ $rg.\textbf{\textit{mes}}_i$ $\leftarrow$ $\sum\limits_{j \in {{\cal N}_i}} {rg.\textbf{\textit{mes}}_{j}}, \ \forall \ i \in rg $ \;     
                }
            $\textbf{\textit{InMap}} = rg.\textbf{\textit{mes}} $ \; 
             /* Return  Instance Map \textbf{\textit{InMap}} */
        \DecMargin{0.25em}
    \end{minipage}
    \end{center}
\end{algorithm}

Note that the GCM used in EFNet is different from the one described above. In the GCM used in EFNet, the initial messages of all the nodes are set as default value of 1, i.e., $tg.\textbf{\textit{mes}}=1$. That means using the displacement field $\hat{\textbf{\textit{F}}}$ to cluster all the nodes in EFNet. Specifically, $\hat{\textbf{\textit{F}}}$ is firstly patched and averaged within each patch to the same size with the input features of GET, and then the processed displacement is used to cluster nodes in GET. The cluster result is finally used to constrain the diffusivity $s_{ij}$ in (\ref{eq7}), that means:
\begin{equation} \label{eq15}
s_{ij}^{(t)}=\left\{\begin{array}{cc}
s_{ij}^{(t)} & \text { if } cls_i = cls_j \\
0 & \text { else }
\end{array}\right.
\end{equation}
where $cls_i$ and $cls_j$ denote the clustering IDs of nodes $i$ and $j$ by GCM, respectively. With constraint of (\ref{eq15}), the message will pass only between the intraclass nodes, which is beneficial for promoting the feature consistency in a given class. 

In the GCM that used for segmenting the adherent glands, the initial messages of all the nodes are determined by the semantic map output by EFNet, i.e., $tg.\textbf{\textit{mes}}=\hat{\textbf{\textit{E}}}=\{0, 1\}$, which allows GCM to refine the clustering result, avoiding the errors caused by displacement field.

\section{Experimental Settings}
\subsection{Dataset and Evaluation}
The GlaS dataset comes from the Gland Challenge of MICCAI 2015 \cite{b1}, including 165 images, 
of which 85 are training set and 80 are test set. Images are downsampled to 256$\times$384 as inputs, and the predictions are upsampled to the original size for evaluation. 
We use three metrics which are widely adopted in the literature on gland instance segmentation \cite{b15}-\cite{b24} as a fair evaluation of model performance, including object-level F1 score (Obj-F1), object-level dice (Obj-Dice), and object-level Hausdorff distance (Obj-HD). The higher Obj-F1 and Obj-Dice indicate the higher accuracy of the model, the lower Obj-HD indicates that the 
detected boundaries of the gland instances are closer to the real ones.

\subsection{Implementation Details}
We use Pytorch and Deep Graph Library (DGL) \cite{b40} to implement DFGET. The GET of EFNet has six stacked GETBlocks, while the GET of DFNet contains 4 stacked GETBlocks. In addition, in the GET of EFNet, the neighborhood of a target node is limited in a 17$\times$17 square area, but in the GET of DFNet, the neighborhood of a target is a circular area with radius 4. In (\ref{eq8}), when we calculate the ground truth displacement field, the neighborhood of a target is also a circular region with radius 5, and the number of iterations is 96. The message passing times for $tg$ and $rg$ of GCM in \textbf{Algorithm 1} are $T_0=2$ and $T_1=8$, respectively.

During training, DFNet is first trained for 400 epochs using the Euclidean distance between the predicted displacement and the true displacement as the loss function, and then EFNet is trained for 300 epochs with MSE and IoU loss as the loss function. Both branches use Adam as optimizer with initial learning rate $lr=0.001$ and $Batchsize=1$. To improve the robustness and generalization ability of the model, we adopt the following data augmentation strategies: 1) Rotating around the center of the image at any angle; 2) Flipping up and down or left and right with a probability of 50\%, respectively; 3) Randomly selecting one of the three geometric transformations (ElasticTransform, GridDistortion, OpticalDistortion); 4) Randomly selecting one of the three color transformations (ColorJitter, RandomGamma, RandomBrightnessContrast); 5) Gaussian blur with a probability of 50\%.

\section{Results}

\begin{table*}[!t] 
    \centering
    \caption{The Quantitative Results of Different Comparison Methods on GlaS Test Set 
    (Results on Part A and Part B Are Averaged).}
    \label{tab_SOTA}
    \renewcommand\arraystretch{1.1}
    \normalsize 
    \begin{tabular}{
        >{\centering\arraybackslash}p{80pt}
        >{\centering\arraybackslash}p{48pt}
        >{\centering\arraybackslash}p{24pt}
        >{\centering\arraybackslash}p{48pt}
        >{\centering\arraybackslash}p{24pt}
        >{\centering\arraybackslash}p{48pt}
        >{\centering\arraybackslash}p{24pt}
        >{\centering\arraybackslash}p{48pt}}
    \hline
    \multirow{2}{*}{Models} & \multicolumn{2}{c}{Obj-F1} & \multicolumn{2}{c}{Obj-Dice} & \multicolumn{2}{c}{Obj-HD} & \multirow{2}{*}{Rank Sum} \\ 
    \cline{2-7}  & Score (\%) & Rank & Score (\%) & Rank & Score (\%) & Rank & ~ \\ \hline
    \textbf{DFGET} & \textbf{94.48} & \textbf{1} & \textbf{93.53} & \textbf{1} & \textbf{32.50} & \textbf{1} & \textbf{1} \\ 
    DIFFormer & 91.37 & 2 & 91.26 & 2 & 48.08 & 2 & 2 \\ 
    U-net & 90.04 & 5 & 90.25 & 5 & 52.16 & 5 & 4 \\ 
    DeepLabV3 & 88.19 & 10 & 90.01 & 7 & 52.82 & 6 & 8 \\ 
    Unet 3+ & 89.81 & 6 & 89.94 & 8 & 52.98 & 7 & 7 \\ 
    UTNet & 89.76 & 9 & 90.26 & 4 & 50.24 & 3 & 5 \\ 
    TCC-MSFCN \cite{b18} & 89.80 & 7 & 89.93 & 9 & 53.20 & 8 & 9 \\ 
    Shape-Aware \cite{b20} & 90.80 & 3 & 89.35 & 10 & 58.76 & 10 & 8 \\ 
    I$^2$CS \cite{b24} & 89.79 & 8 & 90.80 & 3 & 55.40 & 9 & 6 \\ 
    Ta-Net \cite{b41} & 90.50 & 4 & 90.20 & 6 & 50.80 & 4 & 3 \\ \hline
    \end{tabular}
    \vspace{-10pt}
\end{table*}

\subsection{Comparison to SOTA Methods}

To compare the performance of DFGET with different models, we conducted a lot of experiments, 
and the results are shown in Table \ref{tab_SOTA}. Among them, DIFFormer uses message passing in \cite{b35} instead of GETConv. To compare the impact of different network backbones, we use four 
mainstream semantic segmentation networks as comparison models, namely U-net \cite{b9} , DeepLabV3 \cite{b10}, Unet 3+ \cite{b14}, UTNet \cite{b42}. The EFNet branch of DFGET is replaced using the above four semantic segmentation networks and the rest of the settings are kept unchanged. In addition, Table \ref{tab_SOTA} also compares four SOTA models published in recent years, namely TCC-MSFCN \cite{b18}, Shape-Aware \cite{b20}, I$^{2}$CS \cite{b24}, and Ta-Net \cite{b41}. Note that the last four rows in Table \ref{tab_SOTA} directly quote results from the literature.

\begin{figure*}[t]
    \centering
    \includegraphics[width=0.8\textwidth]{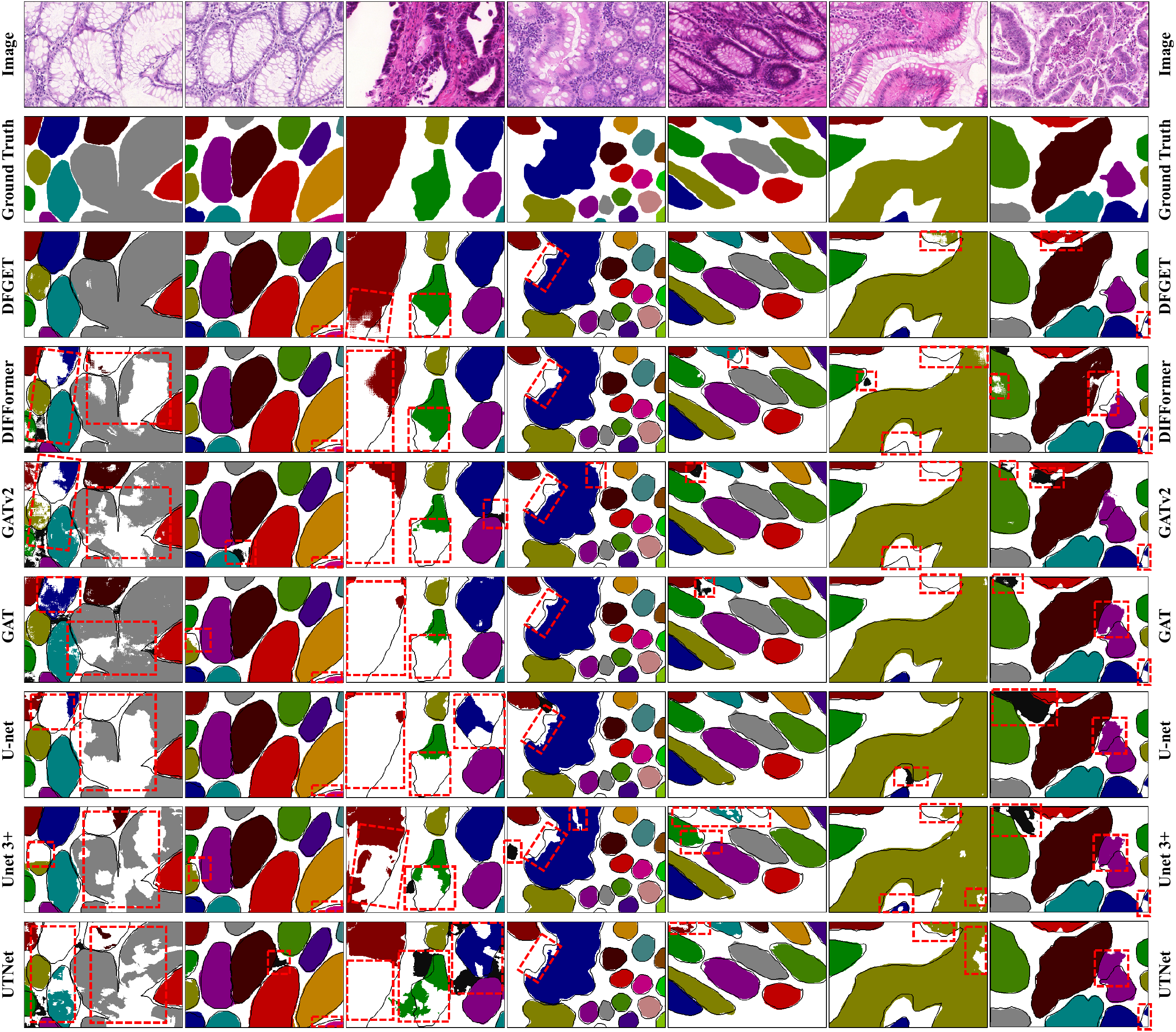}
    \caption{
        Visualization of segmentation results for some challenging samples in 
        the GlaS dataset. The black solid lines in the figure indicate the true 
        instance boundaries, and the red dashed boxes indicate some obvious mis-segmentation.
         }
    \label{fig3}
    \vspace{-10pt}
\end{figure*}

From the Rank Sum of Table \ref{tab_SOTA}, our DFGET gets the best performance (Lower rank means better performance). Specifically, compared to DIFFormer, DFGET improves Obj-F1 by 3.11\%, Obj-Dice by 2.27\%, and decreases Obj-HD by 15.58. Compared to traditional semantic segmentation networks (from U-net to UTNet), DFGET has significant performance gains. For example, compared to U-net, DFGET improves 4.44\% and 3.28\% on Obj-F1 and Obj-Dice, respectively, and decreases 19.66 on Obj-HD. Fig. \ref{fig3} gives the visualization of some difficult samples. The first and second rows show the original image and the ground truth instance labels. The third to ninth rows visualize the segmentation results of our DFGET as well as those of the comparison models. In Fig. \ref{fig3}, the black solid lines indicate the true instance boundaries, and the red dashed boxes indicate some obvious mis-segmentation. We can see that our DFGET performs well on most difficult samples. Even in the most challenging three samples (the first column, the third column, and the seventh column), although DFGET has some mis-segmentations, it is still significantly better than the comparison models. In contrast, comparison models have shown a higher number of false negatives (misclassifying foreground instances as white background) and false positives (misclassifying background regions as colored foreground instances) in several challenging samples. From the above quantitative and qualitative analyses, it is clear that our DFGET significantly outperforms the comparison models.

\subsection{Advantage of GETConv over Other GNNs}
We use different GNNs to replace GETConv and examine the impact of message passing mechanisms. 
The comparison GNNs used include DIFFormer \cite{b35}, GATv2 \cite{b29}, GAT \cite{b26}. As a blank control, 
we exclude GETConv from Fig. \ref{fig1} (b), and get the blank control model IDM (Identity Mapping). 
In order to highlight the role of different message passing mechanisms, the utilization of DF constraint has been temporarily omitted here.
The results are shown in Table \ref{tab_GNNs}, where the metrics of the models using GNN are generally better than that of IDM. Moreover, 
compared to the second-ranked DIFFormer, our GETConv improves Obj-F1 by 0.40\%, Obj-Dice by 1.02\%, 
and reduces Obj-HD by 5.25. These results suggest that GETConv with anisotropic diffusion is more 
expressive for complex glands than the traditional graph attention models with isotropic diffusion. 
The visualization of the segmentation results is shown in rows 3-6 of Fig. \ref{fig3} (DFGET, DIFFormer, GATv2, GAT). 

\begin{table}[!t] 
    \centering
    \caption{Performance of GETConv And Other GNNS on GlaS Test Set (Results on Part A and Part B Are Averaged).}
    \label{tab_GNNs} 
    \renewcommand\arraystretch{1.1}
    \normalsize 
    \begin{tabular}{
        >{\centering\arraybackslash}p{44pt}
        >{\centering\arraybackslash}p{44pt}
        >{\centering\arraybackslash}p{60pt}
        >{\centering\arraybackslash}p{44pt}}
        \hline GNNs & Obj-F1(\%) & Obj-Dice(\%) & Obj-HD \\ 
        \hline \textbf{GETConv} & \textbf{91.77} & \textbf{92.28} & \textbf{42.83} \\  
        DIFFormer & 91.37 & 91.26 & 48.08 \\ 
        GATv2 & 90.49 & 91.30 & 48.14 \\ 
        GAT & 90.85 & 91.17 & 49.71 \\ 
        IDM & 89.21 & 89.51 & 55.37 \\ \hline
    \end{tabular}
    \vspace{-10pt}
\end{table}

\begin{table}[!t] 
    \centering
    \caption{Effect of Displacement Field on Segmentation Performance (Results on Part A and Part B Are Averaged).}
    \label{tab_DF_BD} 
    \renewcommand\arraystretch{1.1}
    \normalsize 
    \begin{tabular}{
        >{\centering\arraybackslash}p{44pt}
        >{\centering\arraybackslash}p{44pt}
        >{\centering\arraybackslash}p{60pt}
        >{\centering\arraybackslash}p{44pt}}
        \hline
        Methods & Obj-F1(\%) & Obj-Dice(\%) & Obj-HD \\ \hline
        \textbf{DFGET} & \textbf{94.48} & \textbf{93.53} & \textbf{32.50} \\ 
        GET & 87.60 & 89.84 & 54.29 \\ 
        BDGET1 & 80.10 & 86.13 & 69.05 \\ 
        BDGET2 & 87.11 & 89.33 & 57.52 \\ \hline
    \end{tabular}
    \vspace{-10pt}
\end{table}

\subsection{Advantage of Displacement Field}
\begin{figure*}[t] 
    \centering
    \includegraphics[width=0.99\textwidth]{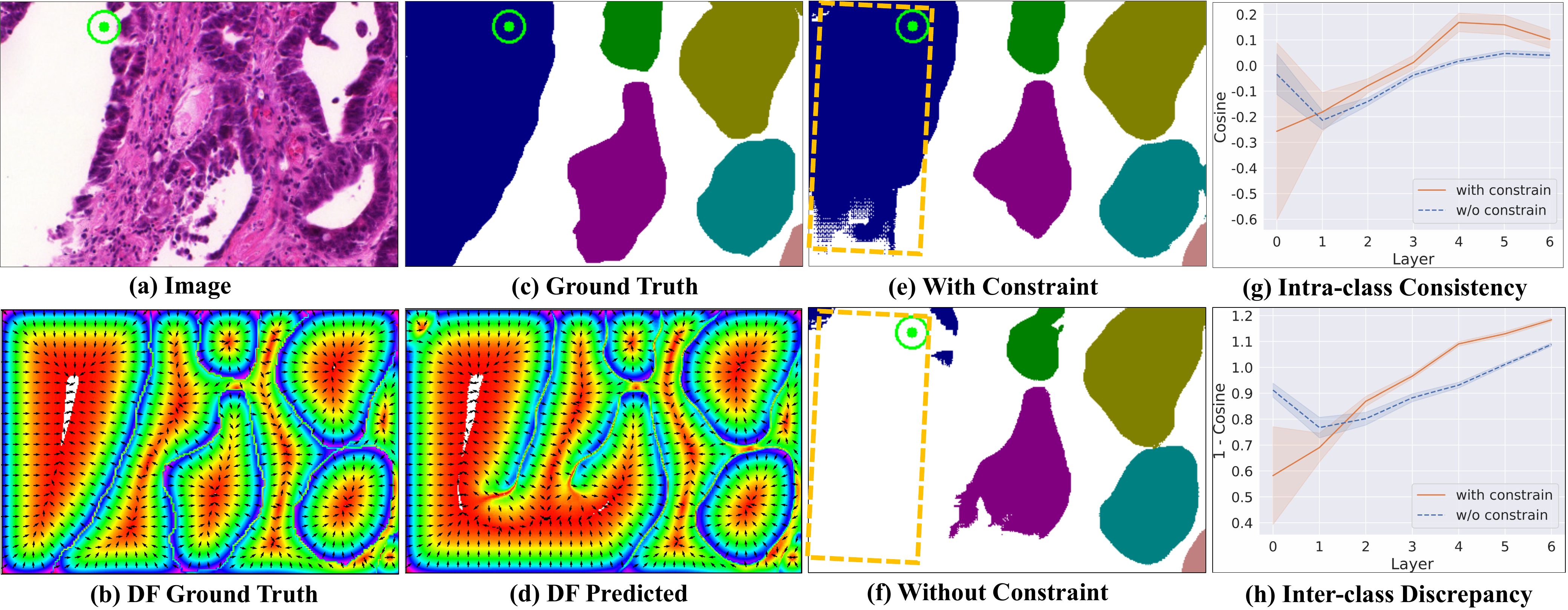}
    \caption{
    Effects of DF constraint. (a) Original image; (b) Ground truth of displacement field; (c) Ground truth of instance map; (d) Displacement field predicted by DFNet; (e) Segmentation result with DF constraint, the three metrics are: Obj\_F1=1.0, Obj\_Dice=0.9348, Obj\_HD=52.61; (f) Segmentation result without DF constraint, the three metrics are: Obj\_F1=0.7692, Obj\_Dice=0.8342, Obj\_HD=160.14; (g) Trend of feature cosine similarity of the target node (center of the green circle) with respect to the foreground, where cosine is on the vertical axis and the layers of GETConvs is on the horizontal axis; (h) Trend of feature discrepancy (1-Cosine) of the target node with respect to the background. 
    }
    \label{fig4}
\end{figure*}

As described in Methods, the GETConvs of EFNet use DF constraint in message passing with (\ref{eq15}), which can enhance the intra-class consistency and inter-class discrepancy of feature representations, thereby improving the output of the energy field network. Fig. \ref{fig4} describes qualitatively and quantitatively the gain of DF constraint for instance segmentation. Without DF constraint, a large area of false negative appears in the yellow dashed box in Fig. \ref{fig4} (f), whereas the corresponding region in Fig. \ref{fig4} (e) can be correctly segmented with DF constraint. In addition, both the intra-class consistency and inter-class discrepancy of features show an overall upward trend, but the extent of the increase is higher for models using DF constraint (Fig. \ref{fig4} (g) and (h)). The quantitative segmentation metrics for the sample in Fig. 4 are shown in captions (e) and (f), where the model with DF constraint outperforms the model without (w/o) DF constraint on two metrics, Obj\_F1 and Obj\_Dice, by 23.08\% and 10.06\%, respectively, and reduces the Obj\_HD by 107.53.

\begin{figure*}[t] 
    \centering
    \includegraphics[width=0.99\textwidth]{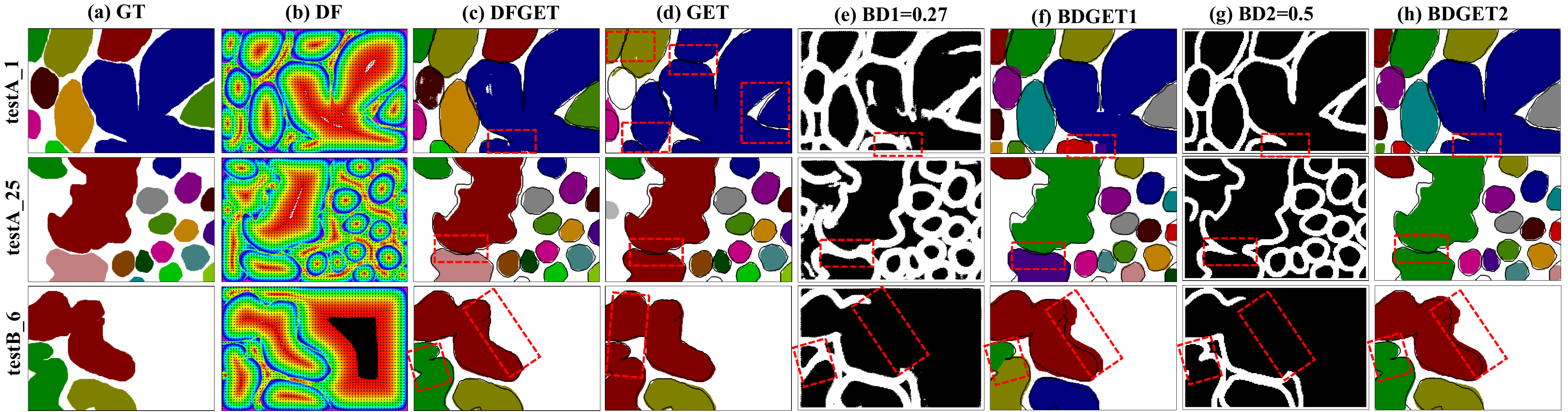}
    \caption{
        Advantages of Displacement Field (DF) over Boundary (BD). (a) GT: Ground 
        truth of instance map; (b) Displacement field output by DFNet; (c) Results 
        of DFGET; (d)Results of GET; (e) Boundary segmentation results at threshold 
        0.27; (f) Results of BDGET1 at threshold 0.27; (g) Boundary segmentation 
        results at threshold 0.50; (h) Results of BDGET2 at threshold 0.50. The black 
        solid lines in (c), (d), (f), and (h) indicate the ground truth instance boundaries.
    }
    \label{fig5}
\end{figure*}

Since DF works through the GCM, we can observe how segmentation performance changes after removing the GCM, thus validating the significance of DF for resolving gland adhesion. To this end, DFGET is degraded to GET by replacing GCM in Fig. \ref{fig1} with OpenCV's morphological opening and connected component detection. Comparing the results in Table \ref{tab_DF_BD}, the segmentation metrics of DFGET are significantly superior than that of GET. Differences between the two model results can be observed in the third and fourth columns of Fig. \ref{fig5}, depicted as (c) DFGET vs. (d) GET. The segmentation results of DFGET are largely consistent with the ground truth annotations, whereas the segmentation by GET without DF exhibits more instances of adhesion. It is evident that, the regions enclosed within the red dashed boxes in Fig. \ref{fig5} (d), which should represent distinct instances of different colors, have been erroneously merged into a single color due to adhesion. 
To compare the advantages and disadvantages of DF with boundary segmentation, DFNet is replaced with a boundary segmentation network which has the same structure as DFNet, only with a different learning objective. The models under two boundary thresholds (BD1=0.27, BD2=0.5) 
are denoted as BDGET1 and BDGET2. As shown in Fig. \ref{fig5}, the results of boundary segmentation are different under different thresholds (see the red dashed boxes 
in (e) and (g)). A smaller threshold (BD1=0.27) makes the boundaries of testA\_25 more 
complete, and the adhered instances in the dashed boxes can be separated. However, a 
small threshold also leads to false positive boundaries in the rest samples, which further leads to incomplete instance segmentation (the red dashed boxes in 
(f) for testA\_1 and testB\_6). Conversely, using a larger threshold can avoid 
false positive boundaries, but it also leads to false negative boundaries and 
adhered instances (see the red boxes in (g) and (h) for testA\_25). It can be seen from Table \ref{tab_DF_BD} and Fig. \ref{fig5} that DF has a more robust and superior performance compared to boundary segmentation.

\section{Discussion}
In this paper, a novel message passing model GET is proposed based on anisotropic diffusion, which is used to update the node information on the graph and solve the problem of limited expressive power of existing graph models. In addition, we introduce a diffusion-based displacement field to drive nodes clustering, which not only constrains message passing but also separates adherent gland instances. By combining GET and displacement field, we obtain DFGET and improve the accuracy of gland segmentation. Numerous experiments demonstrate that the proposed DFGET is more suitable for gland segmentation than comparison models, this section discusses three aspects to analyze the reasons.

\subsection{Anisotropic Diffusion}
We know from the Methods section that GET differs from traditional graph attention models mainly in the diffusion properties of query or key messages. From (7) and Fig.1 (b), it can be seen that the diffusion properties of query message will directly affect the distribution of diffusivity, which in turn affects the semantic message passing between nodes on the graph. 
Therefore, diffusivity distribution map is an important tool to visualize the effect of query-message-diffusion characteristics and analyze the performance of GNNs. 
Fig. \ref{fig6} intuitively demonstrates the difference between the 
traditional graph attention model and our GET. The former has isotropic diffusion of query 
and key messages into the neighborhood, whereas the latter produces anisotropic 
diffusion of queries based on the relative positions of the neighboring nodes. This directly 
leads to the difference in diffusivity distribution between DIFFormer (isotropic) 
and GETConv (anisotropic). As shown in Fig. \ref{fig6}, DIFFormer exhibits a more symmetric diffusivity distribution in the neighborhood compared to GETConv. There are two main reasons for this observation. On the one hand, the diffusion of the query and key messages in DIFFormer is isotropic. On the other hand, DIFFormer calculates the similarity between source and target nodes using inner product, which results in higher isotropy of diffusivity in cases where neighborhood features are highly homogeneous.
In contrast, due to anisotropic diffusion, larger values of GETConv's 
diffusivity are concentrated within the same instance as the target 
node (see Fig. \ref{fig6} (b)). The higher the diffusivity of the source node 
with respect to the target node, the greater its impact on the target 
node in message passing. Thus, with the above analysis, it's clear that 
GETConv can focus the model's attention on the regions that are more 
helpful for the final segmentation.

\subsection{Displacement Field Constraint}
\begin{figure}[!t] 
    \centering
    \includegraphics[width=0.49\textwidth]{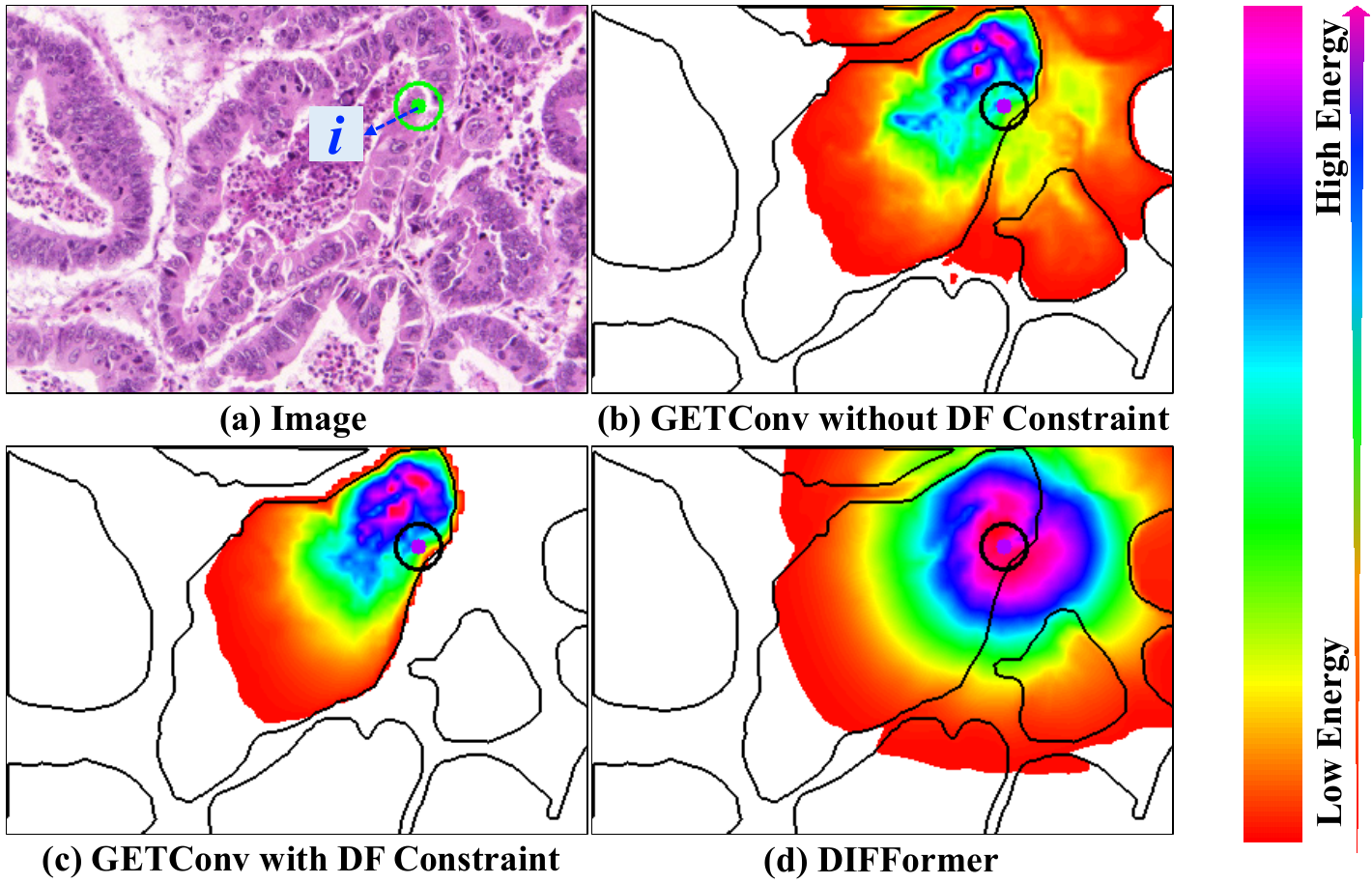}
    \caption{
    Visualization of diffusivity. A node $i$ is selected as the target node 
    (the center of the green or black circle is the position of the target node), 
    and the diffusivity of its 6-order (corresponding to the 6-layer GETBlocks of EFNet) neighborhood on the graph is visualized as a heat map. The boundaries 
    of the instances are indicated by black solid lines in the figure.
    }
    \label{fig6}
    \vspace{-14pt} 
\end{figure}

From Fig. \ref{fig4} (g) and (h), we can find that the DF constraint improves the intra-class consistency and inter-class discrepancy of the output features by GETConv, thus improving the segmentation performance of the model. The underlying mechanism is that DF can cluster nodes belonging to the same instance into one cluster through GCM, and then constrain message passing by formula (\ref{eq15}). As a result, messages are passed between nodes within the same instance, but not between nodes of different instances. Thus, the features of nodes from the same instance become more and more similar, which is conducive to instance segmentation. As in Fig. \ref{fig6} (c), with DF constraint, the diffusivity is concentrated within the same instance, and the neighborhood features that the target node $i$ can obtain mainly come from the nodes of the same instance as $i$. On the contrary, GETConv in Fig. \ref{fig6} (b) has a more scattered distribution of diffusivity due to the lack of DF constraint. Thus, target node $i$ in Fig. \ref{fig6} (b) can 
obtain the features from background and other instances, which leads to lower intra-class consistency and inter-class discrepancy than when using DF constraint. As a result, the classifier of EFNet needs to fit more deeply to distinguish between the target node and background in Fig. \ref{fig6} (b), consequently increasing the risk of overfitting. This also explains why models that use DF constraint outperform models that do not.

\subsection{Robustness of Displacement Field}
We argue that DF can solve the problems of poor boundary segmentation robustness and 
gland adhesion. As shown in Fig. \ref{fig1}, GCM takes DF and EF as input and outputs the 
clustering of nodes. Unlike boundary segmentation, GCM does not need boundary 
thresholds, thus avoiding the uncertainty of boundary thresholds. Moreover, 
$tg$ used for contraction of instances and $rg$ used for recovering the complete instances in GCM are completely reversible, avoiding false positive in the process of restoring complete instances (see testB\_6 in Fig. \ref{fig5} (c), (f), and (h) for comparison between DF and BD). 

In contrast, the models based on boundary segmentation and erosion-dilation postprocessing 
(BDGET1 and BDGET2) require proper boundary threshold. As analyzed in subsection C of the Results section, different boundary thresholds lead to different segmentation results and it is difficult to find a proper threshold that fits all samples. Moreover, since the erosion-dilation operations are not completely reversible, it is difficult to recover the true instance boundaries. For example, the false-negative boundaries within the dashed boxes on the right side of (e) and (g) for testB\_6 in Fig. \ref{fig5} cause the boundaries recovered by dilation operation to exceed the true instance contours (black solid lines), resulting in false-positive instance boundaries in the corresponding dashed boxes of (f) and (h) for testB\_6 in Fig. \ref{fig5}. 

Although DFGET performs well, it has two main limitations: the two-branch architecture increases the computational overhead; and the displacement field is not suitable for dealing with cross-overlapping structures. To apply it to other types of data (such as neuron segmentation) and obtain a general biomedical image instance segmentation model, there are still some key issues to be addressed, such as unifying the displacement field and energy field from the perspective of graph diffusion, and unifying the tasks of the two-branch networks in this paper into node prediction and edge prediction. Thus, we can predict the semantic category on nodes and predict the connection relationship (clustering) on edges, so as to achieve instance segmentation in one-branch graph neural network. Also, based on the connectivity predicted on edges, the nodes can be clustered directly on the graph, thus being able to deal with cross-overlapping structures.

\section{Conclusion}
This paper presents the DFGET framework for gland instance segmentation, 
which is best characterized by the displacement field and anisotropic 
diffusion. To deal with the poor robustness of boundary segmentation 
and gland adhesion in existing models, we derive algorithms to compute 
displacement field from instance labels based on the discrete form of 
diffusion partial differential equation, and use the displacement field 
as a new learning objective. We further propose GCM based on the 
displacement field, which separates adhered instances by displacement-driven clustering.

Meanwhile, the limitations of existing graph neural networks are analyzed from the 
perspective of graph diffusion. Then, the anisotropic GET is proposed, which 
has stronger expressive ability for difficult-to-learn gland instances. Numerous 
experiments and visualization analyses illustrate quantitatively and qualitatively, 
that the DFGET proposed in this paper significantly outperforms the existing models. 
Moreover, the anisotropic diffusion kernel as well as various manipulations of the 
visual nodes on the graph may inspire other medical image segmentation tasks.


\end{document}